\documentclass[10pt,twocolumn,letterpaper]{article}

\usepackage[pagenumbers]{cvpr} % To force page numbers, e.g. for an arXiv version

\usepackage{graphicx}
\usepackage{amsmath}
\usepackage{amssymb}
\usepackage{booktabs}
\usepackage{xcolor}

\usepackage[pagebackref,breaklinks,colorlinks]{hyperref}

\usepackage{bbm}
\usepackage[capitalize]{cleveref}
\crefname{section}{Sec.}{Secs.}
\Crefname{section}{Section}{Sections}
\Crefname{table}{Table}{Tables}
\crefname{table}{Tab.}{Tabs.}

\def\cvprPaperID{11507} % *** Enter the CVPR Paper ID here
\def\confName{CVPR}
\def\confYear{2022}

\newcommand{\Tau}{\mathcal{T}}
\DeclareMathOperator*{\argmax}{arg\,max}

\newcommand{\blu}[1]{{\color{black}{#1}}}
\begin{document}

%%%%%%%%% TITLE - PLEASE UPDATE
\title{Causal Imitative Model for Autonomous Driving}

\author{Mohammad Reza Samsami\\
Sharif University of Tech.\\
% Institution1 address\\
{\tt\small mohammadRezaSamsami76@gmail.com}
% For a paper whose authors are all at the same institution,
% omit the following lines up until the closing ``}''.
% Additional authors and addresses can be added with ``\and'',
% just like the second author.
% To save space, use either the email address or home page, not both
\and
Mohammadhossein Bahari\\
EPFL\\
% First line of institution2 address\\
{\tt\small mohammadhossein.bahari@epfl.ch}
\and
Saber Salehkaleybar\\
Sharif University of Tech.\\
% First line of institution2 address\\
{\tt\small saleh@sharif.edu}
\and
Alexandre Alahi\\
EPFL\\
% First line of institution2 address\\
{\tt\small alexandre.alahi@epfl.ch}
}
\maketitle

%%%%%%%%% ABSTRACT
\begin{abstract}
Imitation learning is a powerful approach for learning autonomous driving policy by leveraging data from expert driver demonstrations. 
However, driving policies trained via imitation learning that neglect the causal structure of expert demonstrations yield two undesirable behaviors: inertia and collision. 
In this paper, we propose Causal Imitative Model (CIM) to address inertia and collision problems. CIM explicitly discovers the causal model and utilizes it to train the policy. Specifically, CIM disentangles the input to a set of latent variables, selects the causal variables, and determines the next position by leveraging the selected variables. 
Our experiments show that our method outperforms previous work in terms of inertia and collision rates. Moreover, thanks to exploiting the causal structure, CIM shrinks the input dimension to only two, hence, can adapt to new environments in a few-shot setting. Code is available at \href{https://github.com/vita-epfl/CIM}{https://github.com/vita-epfl/CIM.}
\end{abstract}

%%%%%%%%% BODY TEXT
\section{Introduction}
\label{sec:intro}
Imitation learning is a desirable approach for learning the driving task as it uses off-line demonstrations of experts \cite{rhinehart2018deep,dosovitskiy2017carla,codevilla2019exploring,filos2020can,Zhang_learnbywatch}. 
Although previous work has achieved successes for learning the task, yet two critical issues exist. First, when the trained agent stops (\eg, at a traffic light), it often stays static. This problem is usually known as ``inertia problem'' and has been observed in \cite{codevilla2019exploring}. Second, the trained agent does not avoid colliding with other cars, known as ``collision problem'', as shown by our experiments.
Although these two problems are critical, they have not received a concrete joint study, to the best of our knowledge.
In this work, we propose a causal framework to address these two problems jointly. It is worth noting that studying inertia and collision problems jointly is essential because one might come up with a solution that resolves one of the problems at the expense of the other one. For instance, a method can speed up the car in some situations to avoid inertia; however, wrong decisions might cause collisions.

\begin{figure}
    \centering
    \includegraphics[width=1\linewidth]{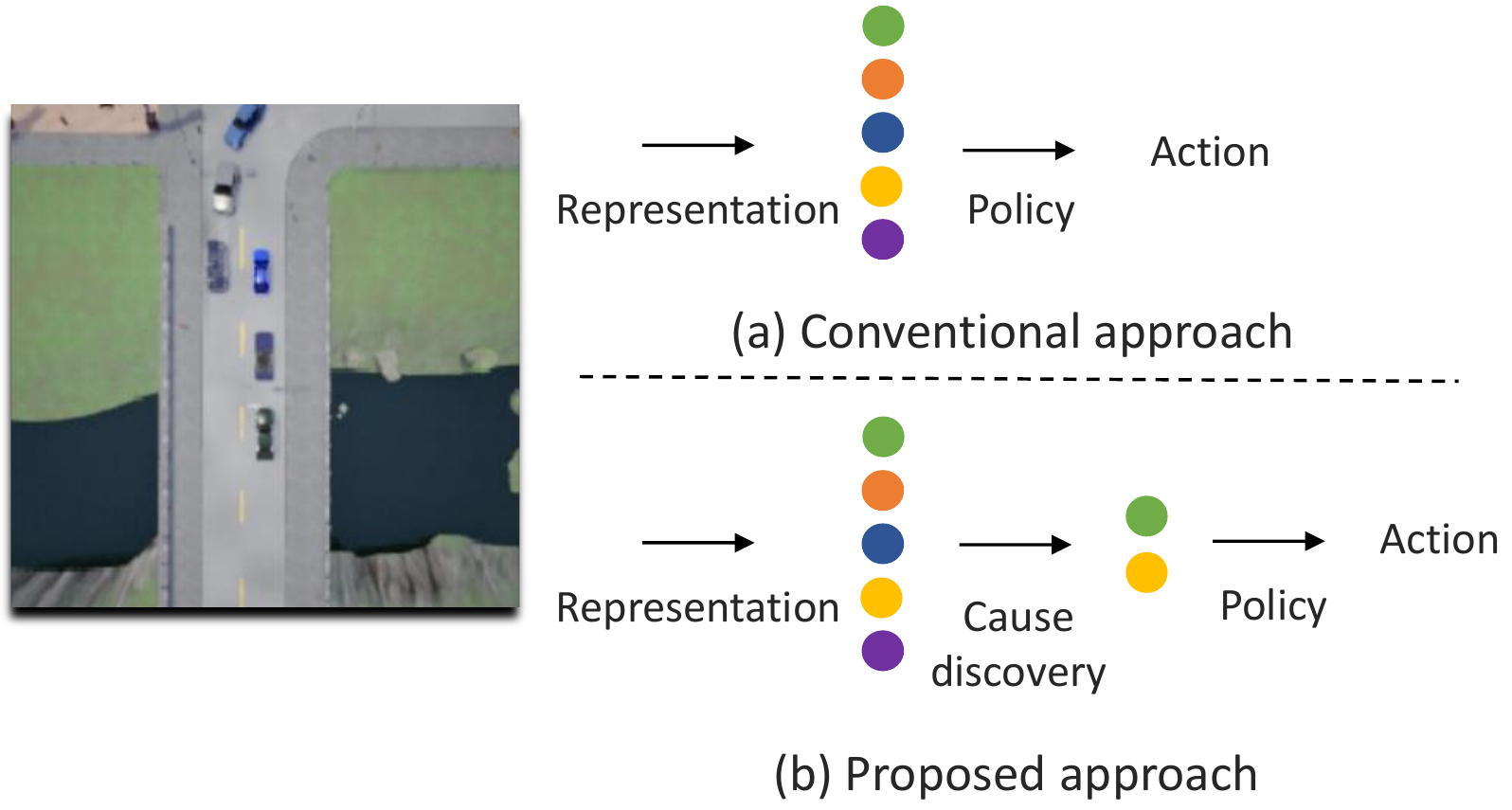}
    \caption{Comparing the conventional approach with the proposed Causal Imitative Model (CIM). Instead of directly learning the policy from input representation (conventional approach), we first detect the causes and then, use them to train the policy (proposed approach).  }
    \label{fig:pull}
\end{figure}

% merge paragraphs

Causality in imitation learning has recently been investigated. Codevilla et al., \cite{codevilla2019exploring} identified learning from spurious correlations rather than real causes as the source of inertia in trained autonomous driving policies.
Haan et al., \cite{Haan2019causal} showed the phenomenon of learning causally-incorrect policies with naive imitation learning. They proposed a causal learning algorithm to mitigate this issue; however, the complexity of the algorithm grows exponentially with the number of latent variables. Hence, it would be prohibitively expensive for inputs in the autonomous driving task. In this paper, we introduce a computationally-feasible causal model for autonomous driving.

We propose \textit{Causal Imitative Model} (CIM) for autonomous driving to mitigate inertia and collision problems. CIM disentangles the input to a set of latent variables. Then, from the latent variables, it selects the causal variables for determining the next position using Granger causality \cite{granger1963economic}. Finally, the policy is trained to find appropriate actions leveraging the causal variables. \cref{fig:pull} visualizes the main idea of CIM. We conduct experiments using CARLA simulator~\cite{dosovitskiy2017carla} and state-of-the-art baselines and demonstrate that CIM reduces inertia and collision problems by more than 75\% and 30\%, respectively. We show that CIM finds meaningful latent variables as the causes of next actions. Specifically, it learns the task employing only two causal latent variables. Therefore, by leveraging a shallow network, it can adapt to new domains with a few samples (few-shot domain adaptation). 
Our main contributions are summarized as follows:
\begin{itemize}
    \item We introduce CIM, an imitative model which learns the autonomous driving policy based on the discovered causes.
    \item We demonstrate that CIM learns meaningful causes and compared to state-of-the-art methods, reduces inertia and collision problems by more than 75\% and 30\%, respectively. It is also able to adapt to new domains with few examples.
    \item We conduct ablation studies to reveal the role of each building block of CIM in the final results. 
\end{itemize}

\section{Related work}
\subsection{Imitation learning for autonomous driving}
Imitation learning, also known as learning from demonstrations, is a powerful and practical framework in scenarios that are too challenging to provide a clear reward signal
%if providing the learner with an acceptable policy is straightforward,
and it has been applied to a wide variety of domains \cite{abbeel2010autonomous, chen2015deepdriving, ratliff2007imitation, ziebart2008maximum}. Specifically, in safety-critical tasks like autonomous driving, imitation learning has gained lots of attention, and its first application in autonomous driving dates back to 1989 in \cite{pomerleau1989alvinn}. Imitation learning through behavior cloning is a form of supervised learning which has been studied in \cite{bojarski2016end, chen2015deepdriving, codevilla2018end,rhinehart2018deep,filos2020can,Zhang_learnbywatch,zhang2021end,chen2020learning}.
Zhang et al., \cite{Zhang_learnbywatch} make use of the demonstrations of other vehicles in a given scene to bring 
data-efficiency into the learning.
Rhinehart et al., \cite{rhinehart2018deep} propose a trajectory density model to plan future trajectories. Their approach benefits from imitation learning in acquiring desirable behavior and exploiting the flexibility of model-based reinforcement learning to adapt to new tasks. This model has been improved in \cite{filos2020can} by using a Bayesian algorithm to do epistemic uncertainty-aware planning, advantageous in making robust decisions in out-of-distribution situations.
While these methods take advantage of behavior cloning,
Codevilla et al., \cite{codevilla2019exploring}
explore the limitation of behavior cloning. They recognize weak performance for rare events due to the bias in datasets, high variance of the learned policy because of stochastic learning and causal confusion due to lack of explicit causal model as main challenges of imitation learning in autonomous driving. Inspired by their work, we propose a causal imitative model to address the lack for a causal model. 

\subsection{Causal structure learning}
Learning causal structures from the observational data is one of the primary problems in statistics and machine learning. Causal relationships among a set of variables are commonly represented by a directed acyclic graph (DAG) where there is a direct link from variable $X$ to variable $Y$ if $X$ is direct cause of $Y$. In the literature of causality, recovering the true causal graph from observation data has been studied in two main settings: random variables and time series. 

In the random variable setting, without considering further assumption on the causal system, one can recover the true causal graph up to a Markov equivalence class by utilizing constraint-based \cite{pearl2009causality,spirtes2000causation} or score-based approaches \cite{meek1997graphical,chickering2002optimal}. In order to uniquely identify the causal graph, it is required to consider additional assumptions on the causal system. For instance, LiNGAM algorithm \cite{shimizu2006linear} can return the true causal graph in linear systems with non-Gaussian exogenous noises. Moreover, under some mild assumptions, it has been shown that the true causal graph can be identified in non-linear system with additive exogenous noises \cite{hoyer2009nonlinear}. 

In the setting of time series, most effort was confined to define the statistical definition of causality such as Granger causality \cite{granger1963economic,granger1969investigating} or transfer entropy \cite{schreiber2000measuring}. In 1960's, Granger proposed a definition of causality between random processes. More specifically, in this definition, random process $X^t$ is a Granger-cause of random process $Y^t$ if knowing the past of $X^t$ up to time $t$, can improve the prediction of future of $Y^t$. Later, Marko proposed another notion of causality between time series based on directed information (DI) \cite{marko1973bidirectional}. More recently, inspired by Ganger causality, it has been shown that the causal structure of a dynamical system can be recovered by computing DI from samples of time series \cite{quinn2015directed}. In \cite{peters2013causal}, TiMINO algorithm has been proposed to recover causal structures between time series when the exogenous noises are additive. In \cite{tank2018neural}, by employing multi-layer perceptrons (MLP) and recurrent neural networks (RNN), a class of methods has been proposed to capture non-linear Granger causality in time series. In \cite{heydari2019adversarial}, two non-linear regression methods have been proposed  to recover causal structures for a wide class of causal systems in both random variable and time series settings. In this work, we utilize Granger causality which is simple but effective enough \cite{granger1963economic} to learn the causes of the speed variable as we will see in Section \ref{sec:interpret}.

\subsection{Causal imitation learning}
\label{sec:causal_imit}
Interest in the intersection of causality and imitation learning has been raised very recently. In \cite{ellias2020cofounders}, the authors focused on the feasibility of imitation from the causal perspective. They explained that when sensory inputs of the expert and the learner vary (\textit{i.e.,} they do not have the same causal model), the implicit assumption that imitating the expert translates into high rewards to the learner is not justified. Therefore, they relaxed this assumption and introduced a graphical criterion determining learner's imitability from demonstration data. 

The need for causal modeling in imitation learning was highlighted in \cite{Haan2019causal,codevilla2019exploring}. Authors in \cite{Haan2019causal} claimed that unless we do not maintain an explicit causal model, true causes cannot be easily identified from spurious correlations. Therefore, policies learned via behavior cloning can suffer from causal confusion phenomenon. They proposed an algorithm to resolve this issue by learning the correct causal model and mapping it to an optimal policy. However, the complexity of algorithm grows exponentially with the number of features. In  \cite{codevilla2019exploring}, the authors observed the existence of inertia problem in autonomous driving policies learned from naively imitating the expert. They recognized the main source of the problem as the tendency of the learned policy towards spurious correlations instead of real causes. Moreover, they proposed to use a speed prediction branch as an auxiliary loss to remedy the problem. In contrast, we propose a causal learning method to address the inertia problem.

The structure of the rest of the paper is as follows: In \cref{sec:formulation}, we formulate the problem of imitation learning in self-driving cars. In \cref{sec:method}, we present CIM method and explain its different parts. We provide experimental results for CIM in \cref{sec:experiments} and compare it with previous work. Finally, we conclude the paper in \cref{sec:conclusion}.

\section{Problem formulation}
\label{sec:formulation}
In this section, we formalize our imitation learning problem and introduce some definitions. In an imitation learning setting, the learner utilizes demonstrations from an expert, \eg, a human or a rule-based algorithm, to learn a policy for performing a given task. 

In each time step $t$, the agent has access to a high-dimensional observation of the scene around itself denoted by $O^t$, paired with expert future trajectory $\Tau^t$. The observations include the information of the static parts of the scene, known as the context, and the dynamic agents in the scene.  
The expert policy generates the trajectories, $\Tau \sim \pi(.|O)$, and we aim to approximate $\pi$ by using ($O^t$, $\Tau^t$) tuples. The simplest form of imitation learning is behavior cloning, which reduces a policy learning problem to a supervised learning problem. In this approach, the agent learns to map $\pi$ from $O$ to $\Tau$ directly. We also assume that the agent has access to an inverse dynamic model as a controller. 
Controller receives the current and subsequent positions of the car and offers low-level control commands $A^t$ comprising of steering, throttle, and brake. The agent executes $A^t$ which updates the environment and provides the next observation.

\subsection{The underlying causal mechanism of expert's demonstrations}
\label{sec:expert_mech}

Inspired by the causal model in \cite{Haan2019causal}, we hypothesize that the underlying causal mechanism of expert demonstrations is a sequential decision-making process. Every observation $O^t$ is disentangled into some generative factors $Z^t=[Z^t_1, Z^t_2,..., Z^t_k]$, each representing some  information regarding the state of the world where $k$ is the number of generative factors. For instance, one factor can represent the presence of cars in front or weather condition. We consider $Z^t$ as the state at time step $t$. To plan the future trajectory $\Tau^t$, first, the expert infers the speed for the next timestep $S_\mathrm{forward}^t$ based on the current state $Z^t$.
Then, the expert uses the state and the inferred speed, to induce a trajectory. Note that only some of the factors in the state directly influence the speed variable $S_\mathrm{forward}^t$ and the rest are nuisance variables. We consider the influential factors as the causes (or parent set) of the speed.

\section{Method}
\label{sec:method}

One of the main challenges of autonomous driving is to be able to interact with other agents. If an autonomous driving policy does not account for the interaction correctly, it leads to collisions or inertia. 
We present \textit{Causal Imitative Model} (CIM), a method based on the causal mechanism of expert's demonstrations in \cref{sec:expert_mech} to learn the expert policy. CIM consists of two main parts: 

\begin{itemize}
\item \textbf{An imitative model} that deals with the static context-related navigation in the scene. It estimates the density of the distribution over future expert positions $q(\hat{\Tau} | O ; \theta)$. For this part, we employ an off-the-shelf imitative model and train it using maximum likelihood estimation:
$\theta^* = \underset{\theta}{\argmax}\ \mathbbm{E} \big[\log q(\hat{\Tau} | O ; \theta) \big]$.

\item \textbf{A causal pipeline} that estimates the desired speed and has three consecutive parts. First, there is a \textit{Perception model} that embeds observations into a disentangled representation. Then, the \textit{Causal speed predictor} identifies the direct causes of the speed and finds the next step speed using them.
Finally, the Controller adjusts the imitative model's output by utilizing the predicted speed value. We will describe this pipeline in more detail.
\end{itemize}

\begin{figure*}[t]
\centering
\includegraphics[width=\linewidth]{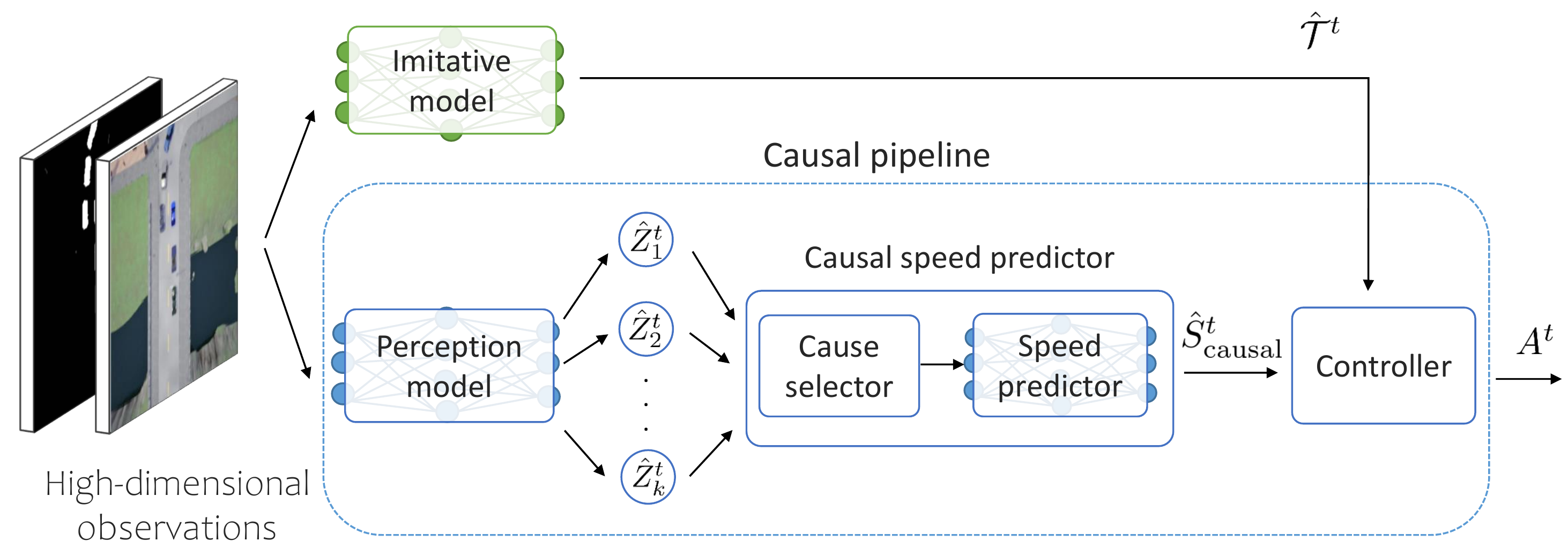}
\caption{Illustration of our method. Imitative model is responsible for providing a static context-based trajectory. CIM deals with the interaction between vehicles, thus, finds the speed of the vehicle. It first disentangles the input into latent variables by \textit{Perception model}. Then, \textit{Cause selector} keeps the causes of the speed. These causes are processed by \textit{Speed predictor} to find the speed. Finally, \textit{Controller} merges the trajectory and the speed and provides control signals. }
\label{fig:method}
\end{figure*}

\subsection{Perception model}

At each time step, we get an observation $O^t$ from the environment. As discussed in section \cref{sec:expert_mech}, the expert makes decisions based on the disentangled generative factors $Z^t$ of the observation. Therefore, the role of the \textit{Perception model} is to learn latent variables $\hat{Z}^t$ as a disentangled abstract representation of observed input $O^t$. In a desired disentangled representation, each latent variable in $\hat{Z}^t$ is sensitive to changes in one generative factor while being relatively invariant to other factors \cite{higgins2018towards}. Variational autoencoders (VAE) \cite{kingma2013auto} are commonly employed to obtain disentangled factors from observations \cite{higgins2016beta, kim2018disentangling, chen2018isolating}. We implemented and trained a $\beta$-VAE \cite{higgins2016beta} as our \textit{Perception model} to compress the observations into a disentangled representation of $128$ variables.

\subsection{Causal speed predictor} \label{sec:predictor}
Having learned the disentangled representation, we need to identify causes of the next step speed $S^t_\mathrm{forward}$ from learned latent variables $\hat{Z}^t$. To this end, we consider time series of $\hat{Z}_i, i=1,\cdots,k$ and take speed as the target variable.
Here, we use the notion of Granger causality \cite{granger1963economic}. In particular, we say $X^t$ is a Granger-cause of $Y^t$ if knowing the past of $X^t$ aids in predicting the future of $Y^t$. More specifically, let $\sigma_{Y}(h|\Omega^t)$ be the mean square error (MSE) of the $h$-step predictor of $Y^t$ at time $t$ given information $\Omega^t$. $X^t$ is said to be a Granger-cause of $Y^t$ if:
\begin{equation}
    \exists h>0, \mbox{ $s.t.$ } \sigma_Y(h|\Omega^t)< \sigma_Y(h|\Omega^t\backslash \{X^i\}^t_{i=0}),
\end{equation}
where $\Omega^t$ includes all the information in the universe related to the past or present of $Y^t$. 

It can be shown that one can identify the true causes of a target variable using Granger causality if we are observing all the variables that may have influences on it\footnote{There are some other mild assumptions that should be taken into account. Please check Theorem 10.3 in \cite{peters2017elements} for more details.}. This assumption is commonly called ``causal sufficiency'' assumption. By considering this assumption, we are able to perform Granger causality tests in order to recover the causes of speed. 

We consider the selected variables from Granger causality tests as the true causes of speed and design \textit{Speed predictor} based on them. \blu{To select these variables, we performed a hypothesis testing where the null hypothesis is that the $\hat{Z}_i$ time series does not Granger cause the ``speed'' time series. We reject the null hypothesis that $\hat{Z}_i$ does not Granger cause speed if the $p$-values are below a threshold.} \blu{Despite its simplicity, Granger causality works fairly well in our setting and identifies the interpretable cause variables (as we see in \cref{sec:interpret}). One might use other techniques but we believe that the selected cause variables would be the same in our setting.}

Similar to the previous work \cite{ha2018worldmodels}, we intentionally train \textit{Speed predictor} independently of the \textit{Perception model} to have latent variables independent of the next block and reside representation complexity in \textit{Perception model}. 
Thanks to our approach, we came up with two causal variables among $128$ latent variables. Our experiments shows that it suffices to consider a shallow network for the predictor fed by the concatenated last three steps of cause variables to predict the speed. This is beneficial in the few-shot adaptation setting. In the experiments, we will look into this further.

\subsection{Controller}

\blu{The trajectory planned by the Imitative model and the speed predicted by Causal speed predictor are fed to the \textit{Controller}. \textit{Controller} then computes the forward speed error, which is the difference between the current and predicted forward speeds. Based on the speed error, the \textit{Controller} uses the control as throttle or brake. Finally, the agent executes the calculated control commands, causing a transition in the environment.
}

\blu{More specifically, suppose at time $t$, the causal predictor returns $\hat{S}_\mathrm{causal}^t$. The \textit{Controller} computes the forward speed error, $e_s^t = \hat{S}_\mathrm{causal}^t - S_\mathrm{current}$ where $S_\mathrm{current}$ is the current speed of the car, and then it computes accelerator control $U_s^t$ using $e_s^t$ \cite{rhinehart2018deep}. At last, it calculates the brake and throttle values as follows:}
\begin{equation}
    \mathrm{throttle} \leftarrow \mathbbm{1}(e_s^t > 0)\cdot U_s^t,
    \mathrm{brake} \leftarrow  \mathbbm{1}(e_s^t \leq 0)\cdot U_s^t
\end{equation}

\section{Experiments}
\label{sec:experiments}
In this section, we evaluate the performance of CIM under multiple settings. We conduct experiments to address the following points: 

\begin{enumerate}
    \item Does the proposed approach reduce collision and inertia events?
    \item Does our method benefit from merits of causal reasoning such as few-shot domain adaptation?
    \item How does each part of the method contribute to the results? 
\end{enumerate}

\subsection{Dataset}
We evaluate the methods using CARLA urban driving simulator \cite{dosovitskiy2017carla}, which is a standard platform for research in autonomous driving and has enabled many researchers to train policies in a dynamic urban environment. CARLA provides various environments with different towns and number of cars. We collected data from \texttt{Town01} and split it into train and test data. We also collect a second test data from \texttt{Town02} as a different environment with distributional shift.
In the experiments, we specified the town, start and destination points, and the number of cars and pedestrians in every task. Any other initial configuration, like locations of other cars, was initialized randomly. We executed three trials per task to reduce the possible influence of randomness in initial states.
%described in detail in supplementary materials.
All the experiments were performed on Nvidia RTX 2080 GPU. 

\subsection{Performance measures}
To study the extent our method is effective in mitigating the inertia and collision problems, we introduce various metrics to measure the percentage of unsuccessful attempts based on inertia and collisions:
\begin{itemize}
    \item \textbf{Inertia rate:} The percentage of navigation tasks that failed due to the inertia problem. Motivated by \cite{codevilla2019exploring}, we consider a task as a failed one due to the inertia problem if the agent has almost zero speed for at least 15 seconds before the timeout.

    \item \textbf{Collision rate:} The percentage of navigation tasks ending up in an accident with another car.

    \item \textbf{Error rate:} Sum of collision rate and inertia rate. Comparing methods based on both inertia and collision rates is challenging as there is a trade-off between these two metrics; to reduce inertia rate, the agent applies more throttle leading to higher speed values; consequently, colliding with the other cars is more likely. 

Therefore, we need to aggregate collision and inertia rates into one metric. In order to give them equal importance, we considered the sum of them as the error rate to quantify the total failures occur in the tasks. We will use this metric to compare different methods throughout the experiments.

\end{itemize}

\subsection{Baselines}
To compare CIM's performance with previous work, we considered two state-of-the-art models, Deep Imitative Model (\textbf{DIM}) \cite{rhinehart2018deep} and Robust Imitative Planning (\textbf{RIP}) \cite{filos2020can}. RIP is an uncertainty-aware planning method based on deep ensembles proposed to tackle out-of-distribution scenarios. In addition, we report a na\"ive baseline that always follows DIM trajectory at constant speed (\textbf{CS}). The upper bound on the collision rate is given by this baseline. 

As ablation studies, we implemented the following baselines; \textbf{CIM-MLP}, which is the same as CIM except that instead of discovering the causes of the speed variable in the time series, it feeds all latent variables into \textit{Speed predictor}; \textbf{CIM-entangled} which is identical to CIM except it learns entangled representation by %setting $\beta=1$ in training $\beta$-VAE.
replacing $\beta$-VAE with the original VAE.

\subsection{Implementation details}\label{sec:pre}
\blu{As stated in \cref{sec:method}, our method is capable of working with any imitative model trained by maximum likelihood estimation. Hence, the process of future trajectory planning is as abstract as $\theta^* = \underset{\theta}{\argmax}\ \mathbbm{E} \big[\log q(\hat{\Tau} | O ; \theta) \big]$ in general. 
In this work, we leverage DIM as the imitative model. We trained DIM with the same settings described in \cite{rhinehart2018deep}.}
We built our work on \textit{OATomobile} to use the same framework as \cite{filos2020can}. It is a framework for autonomous driving research which wraps CARLA in OpenAI gym environments. To collect the demonstrations, we executed a rule-based autopilot expert implemented by \cite{filos2020can} in \texttt{Town01}.

\subsubsection{Input representation}

\blu{Similar to the previous works \cite{bansal2018chauffeurnet,Zhang_learnbywatch,chen2020learning,zhao2019sam,schulter2018learning,wang2019monocular}, we use bird-eye view images as the input to the perception model to account for the static context of the scene. Indeed, the bird-eye view is a format capable of combining all sensory inputs into a unified representation}\footnote{\blu{The traffic light is an event which may have an impact on the ego's car speed. Nonetheless, because our experiments were conducted in busy cities, we observed that usually near to the traffic lights, at least one car is in front of the ego vehicle. Thus, explicitly considering the traffic light state has negligible impact on our results.}}.

\begin{figure}[h]
    \centering
    \subfloat[\centering Bird-view observation ]{{\includegraphics[width=3.75cm]{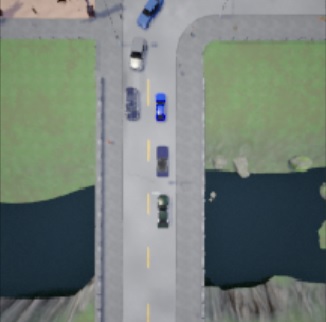} }}%
    \quad
    \subfloat[\centering Other agents representation]{{\includegraphics[width=3.75cm]{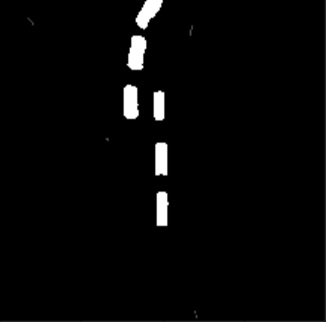} }}%
    \caption{We use the bird-view observation to represent the context of the scene and use the rendered oriented bounding boxes to represent other agents in the scene.}% in order to speed feed the perception model.}%
    \label{fig:preprocess}%
\end{figure}

The status of other agents in the scene can be represented by their 3D locations \cite{Zhang_learnbywatch} or by rendering oriented bounding boxes \cite{zhang2021end,chen2020learning,bansal2018chauffeurnet}. We opt for the latter format to represent the input $O^t$ as it conveniently manages variable agent numbers. Both inputs are shown in \cref{fig:preprocess}.

\subsubsection{Perception model}

For CIM and CIM-MLP, $\beta$ was set to $6$. Adjusting larger value ($\beta > 6$) applies intense pressure on the latent bottleneck and limits the representation capacity, preventing us from analyzing the characteristics of causes of speed. We set $\beta = 1$ for CIM-entangled corresponding to the original VAE \cite{kingma2013auto}.

We first resize each observation to $64 \times 64$ pixels. The model takes in this $64 \times 64 \times 1$ input tensor (it is a gray-scale image) and encodes it into vectors $\mu$ and $\sigma$, each with size $128$. After sampling the latent vector $Z$, it is passed through the decoder layers to reconstruct the observation.

Specifically, the resized image is pass through a sequence of 4 convolution layers with 32 filters each (except the last one that contains 64 filters), kernel size of $4 \times 4$, stride of 2 and padding of 1. The output of the last convolution layer is fed into 3 fully connected networks of 256 units.

In the decoder, the latent vector $Z$ is passed through 3 of fully connected layers (256, 256, $64\times4\times4$ units) followed by 4 of deconvolution layers (with same stride and padding to convolution layers and 32, 32, 32, 1 filters) used to decode and reconstruct the image. A leaky rectified linear unit (Leaky ReLU; \cite{Maas2013RectifierNI}) is used after each layer except the last fully connected layer in the encoder.

\subsubsection{Speed predictor}
We design two different architectures to predict the next speed; each has an encoding layer and a predictor. A fully-connected encoder for each input variable in the encoding layer takes the concatenated last three steps of that variable and encodes it to the hidden representation. For CIM and CIM-entangled, each encoder is a $3 \times 16$ linear model. Moreover, we set every encoder as a $3 \times 8$ linear model for CIM-MLP. All of them have ReLU non-linearity. The predictor is a fully-connected network fed by the concatenated hidden representations and predicts the speed of the next step.

\begin{figure*}[h]
  \centering
  \begin{subfigure}{\linewidth}
  \centering
    \includegraphics[width=0.95\linewidth]{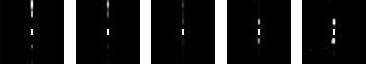}
    \caption{Traversals of the first factor which changes the location of the front and the back cars}
    \vspace*{3mm}
  \end{subfigure}
  \begin{subfigure}{\linewidth}
  \centering
    \includegraphics[width=0.95\linewidth]{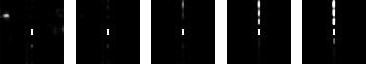}
    \caption{Traversals of the second factor which changes the number of the front cars}
  \end{subfigure}
  \caption{The result of traversal of two cause variables. Each image demonstrates the result of traversing a cause variable. To create these images, we first infer the latent representation. Next, we traverse a single latent variable while keeping the other latent variables unchanged. The depicted results indicate that these two variables are meaningful and essential for interacting with other vehicles for inferring ego car's speed.}
  \label{fig:traversal}
\end{figure*}

\subsection{Interpreting the discovered cause variables}
\label{sec:interpret}

Among $128$ disentangled latent variables, some of them are causes of speed and the rest are nuisances. 
After applying Granger causality, we found that only two latent variables have statistically significant effect on the speed. By applying tools from \cite{locatello2019challenging}, we analyzed the characteristics of these two variables. In particular, we plotted latent space's traversals by changing each of these two variables' values one at a time in the range of $[-3,3]$ while keeping all other latent variables fixed to their inferred values. We observed that traversal of one of the found variables produces \textit{variations in the location of the front and the back cars} to the ego car, and the other one changes \textit{the number of the front cars}, as depicted in \cref{fig:traversal}. Due to an imperfect disentanglement, these two variables, to some extent, represent a combination of some generative factors like the existence of the front car or changing the status of the car behind. However, they are both related to the status of the front car, indicating that a meaningful cause for the speed of ego's car has been discovered.

\subsection{Results}
To address question \textbf{(1)} regarding improvements in inertia and collision rates, we evaluated CIM, DIM, RIP, and CS on 100 tasks in \texttt{Town01}. To compare the methods in an environment with distributional shift, we also assessed them on 100 collected tasks from \texttt{Town02}. 
The results are reported in \cref{tab:res1}. 
Our method has a remarkably better performance than other baselines in all metrics. In \texttt{Town01}, CIM has almost 69\% and 74\% less error rate compared to DIM and RIP, respectively. In \texttt{Town02}, also, CIM could reduce the error rate of DIM and RIP by 49\% and 54\%. The results show the noticeable impact of our approach in reducing the error rate in both in-distribution and out-of-distribution data.

\begin{table*}[h]
  \centering
  \begin{tabular}{lcccccc}
    \toprule
    &
    \multicolumn{3}{c}{\texttt{Town01}}& 
    \multicolumn{3}{c}{\texttt{Town02}} 
    \\
    \cmidrule(r){2-4}
    \cmidrule(r){5-7}
    Methods     & \textbf{Error} $\downarrow$     & Inertia $\downarrow$   & Collision $\downarrow$
    & \textbf{Error} $\downarrow$     & Inertia $\downarrow$   & Collision $\downarrow$  
    \\
    \cmidrule(r){1-1}
    \cmidrule(r){2-4}
    \cmidrule(r){5-7}
    CS & 71\% & 0\% & 71\% & 71\% & 0\% & 71\% \\
    DIM \cite{rhinehart2018deep} & 65\%  & 40\% & 25\% 
    &   74\% & 30\% & 44\% 
    \\
    RIP \cite{filos2020can} & 77\%  & 42\% & 35\% 
    &    83\% & 43\% & 40\% 
    \\
    CIM (ours) & 20\%  & 6\% & 14\% 
    &   38\% & 7\% & 31\% 
    \\
    \bottomrule
  \end{tabular}
  \caption{Comparison of different autonomous driving methods on CARLA’s navigation tasks in \texttt{Town01} and \texttt{Town02}. All of these baselines are performed on 100 different tasks in every town; each contains a specified start and destination points.}
  \vspace*{2mm}
  \label{tab:res1}
\end{table*}

\subsubsection{Ablation studies}
In this part, we report ablation studies to answer questions \textbf{(2, 3)}. More specifically, we want to study the importance of different parts of CIM. Additionally, we want to investigate if CIM benefits from causal learning in the setting of few-shot domain adaptation.
To this end, we assessed CIM and its two variants CIM-MLP, and CIM-entangled on \texttt{Town01} and \texttt{Town02}. The results are provided in \cref{tab:res2}. We discuss the results in the next paragraphs.

\begin{table*}[t]
  \centering
  \begin{tabular}{lcccccc}
    \toprule
    &
    \multicolumn{3}{c}{\texttt{Town01}}& 
    \multicolumn{3}{c}{\texttt{Town02}} 
    \\
    \cmidrule(r){2-4}
    \cmidrule(r){5-7}
    Methods     & \textbf{Error} $\downarrow$     & Inertia $\downarrow$   & Collision $\downarrow$  
    & \textbf{Error} $\downarrow$     & Inertia $\downarrow$   & Collision $\downarrow$  
    \\
    \cmidrule(r){1-1}
    \cmidrule(r){2-4}
    \cmidrule(r){5-7}
    CIM & 20\%  & 6\% & 14\% 
    &   38\% & 7\% & 31\% 
    \\
    CIM-MLP & 22\% & 4\% & 18\% 
    &   42\% & 4\% & 38\%
    \\
    CIM-entagled & 81\%  & 77\% & 4\% 
    &    78\% & 68\% & 12\% 
    \\
    \bottomrule
  \end{tabular}
  \caption{Assessing the effect of components in CIM. CIM-MLP lacks the causal structure learning component used in CIM and instead feeds all latent variables to the speed predictor. Also, disentanglement is absent in CIM-entangled.}
  \label{tab:res2}
\end{table*}

\noindent\textbf{Effectiveness of disentangled representation.}
In order to be able to separate causal and nuisance factors, CIM disentangles input to disentangled latent factors. Otherwise, every latent dimension possibly captures both nuisance variables and causes; consequently, discovering the true causes becomes infeasible.

To investigate the impact of disentanglement, we evaluated the performance of CIM-entangled. As shown in \cref{tab:res2}, \blu{CIM-entangled exhibits poor performance (the error rate is even worse than DIM), indicating that disentanglement plays a vital role.} The low collision rate of CIM-entangled is due to its high inertia rate, which means that the agent is stuck in most cases.

Note that the examination of this ablation demonstrates that although the perception model's inputs have low information, 
a typical vision approach
is unable to learn the interaction admissibly. This also again stresses the challenge of learning the interactions from observations.

\noindent\textbf{Explicit causal discovery and few-shot setting.} 
\textit{Perception model} disentangles input to $128$ latent variables. Then, \textit{Cause selector} in CIM selects causal latent variables by applying causal structure learning algorithms mentioned in \cref{sec:predictor}. 
In other words, we consider the variables that has direct influence on the speed and leave nuisance variables.
To show the advantage of this approach, we compared CIM with CIM-MLP. CIM-MLP learns to predict the speed using all $128$ latent variables with more complex network than CIM and it does not explicitly learn the causal structure.
\blu{According to \cref{tab:res2}, CIM has slightly better results than CIM-MLP. In order to further study the effectiveness of the trained models, we consider the task of few-shot domain adaptation on a new domain.}

Online adaptation of models to new environments is a valuable feature for practical use. Desirably, this adaptation should be achieved with small number of samples. Hence, to study both the capability of models in domain adaptation and their performance in low-data regime, we perform few-shot domain adaptation. 
In this experiment, first, we fully trained speed predictors CIM and CIM-MLP in \texttt{Town01}, and then, tuned them with a low number of samples (100 samples) from another town with a different structure. We selected \texttt{Town03} for the test town
as in contrast with \texttt{Town01}, it is the most complex town in Carla. \blu{Note that while the differences are more obvious in the static context, the dynamic agents distribution and the interaction between agents are also significantly changed.
For instance, in addition to roundabouts, 5-lane junctions and novel 3-way intersections might lead to distinct interaction situations.}

\cref{tab:res4} reports the performance of models and shows that CIM outperforms CIM-MLP in all metrics. This indicates CIM-MLP requires more samples to perform as good as CIM, showing the advantage of the employed causal structure learning.

\begin{table}[h]
  \centering
  \begin{tabular}{lcccc}
    \toprule
    Methods     & \textbf{Error} $\downarrow$     & Inertia $\downarrow$   & Collision $\downarrow$ 
    \\
    \midrule
    CIM-MLP & 38\% & 22\% & 16\% 
    \\
    CIM &   29\% & 20\% & 9\% 
    \\
    \bottomrule
  \end{tabular}
  \caption{Evaluating the performance of CIM and CIM-MLP in few-shot domain adaptation in \texttt{Town03}.}
  \label{tab:res4}
\end{table}

\section{Limitations and future works}

Our work has some limitations which can be addressed in the future work:

\begin{itemize}
    \item The results in \cref{sec:interpret} showed that one of the latent variables selected as causes does not capture a single generative factor. Therefore, the performance of the disentanglement method is one of the work's limitations. Better disentanglement approaches may lead to more disentangled latent variables and potentially better quantitative results.
    
    \item Our work focuses on learning a policy to causally learn the interaction of agents, which reduces the inertia and collision problems. Still, it does not address the scene navigation problem. Future work could extend the methodology to learn scene navigation in a casual approach and merge with the current work.
\end{itemize}

\section{Conclusions}

\label{sec:conclusion}
In this paper, we proposed Causal Imitative Model (CIM) to learn the autonomous driving policy. CIM disentangles the input into latent variables, identifies the causal variables and leverages them to learn a causal policy. This approach addresses the inertia and collision problems which exist in the previous imitation learning works. Experimental results in the CARLA driving simulator validated that our solution successfully employs causally related concepts, has a better performance than state-of-the-art methods in terms of inertia and collision problems, and can adapt to new environments by observing few data points from each of them.

{\small
\bibliographystyle{ieee_fullname}
\bibliography{egbib}
}

\end{document}